\documentclass{article}

\usepackage{PRIMEarxiv}

\usepackage[utf8]{inputenc} 
\usepackage[T1]{fontenc}    
\usepackage{hyperref}       
\usepackage{url}            
\usepackage{booktabs}       
\usepackage{amsfonts}       
\usepackage{nicefrac}       
\usepackage{microtype}      
\usepackage{lipsum}
\usepackage{fancyhdr}       
\usepackage{graphicx}       

\graphicspath{{media/}}     
\usepackage{epstopdf}
\usepackage[pdf]{pstricks}
\usepackage{graphicx, caption, subcaption}
\usepackage{tikz-cd}

\DeclareGraphicsExtensions{.eps}
\usepackage{multirow}
\usepackage{booktabs}

\usepackage{amsmath}

\DeclareMathOperator*{\argmin}{arg\,min}

\title{Looking For A Match: Self-supervised Clustering For Automatic Doubt Matching In {\it e}-learning Platforms 
}

\author{
  Vedant Sandeep Joshi, Sivanagaraja Tatinati \\
  Vedantu Innovations Pvt. Ltd., \\
  Bangalore, India \\
  \texttt{\{vedant.joshi, tatinati.sivanagaraja\}@vedantu.com} \\
    \And
  Yubo Wang \\
  School of Life Science and Technology, \\
  Xidian University,  \\
  Xi-an, China\\
  \texttt{ybwang@xidian.edu.cn} \\
}

\begin{document}
\maketitle

\begin{abstract}
Recently, {\it e-}learning platforms have grown as a place where students can post doubts (as a snap taken with smart phones) and get them resolved in minutes. However, the significant increase in the number of student-posted doubts  with high variance in quality on these platforms not only presents challenges for teacher's navigation to address them but also increases the resolution time per doubt. Both are not acceptable, as high doubt resolution time hinders the students learning progress. This necessitates ways to automatically identify if there exists a similar doubt in repository and then serve it to the teacher as the plausible solution to validate and communicate with the student. Supervised learning techniques (like Siamese architecture) require labels to identify the matches, which is not feasible as labels are scarce and expensive. In this work, we, thus, developed a label-agnostic doubt matching paradigm based on the representations learnt via self-supervised technique. Building on prior theoretical insights of BYOL (bootstrap your own latent space), we propose custom BYOL which combines domain-specific augmentation with contrastive objective over a varied set of appropriately constructed data views. Results highlighted that, custom BYOL improves the top-1 matching accuracy by approximately 6\% and 5\% as compared to both BYOL and supervised learning instances, respectively. We further show that, both BYOL based learning instances performs either on par or better than human labeling. 
\end{abstract}


\section{Introduction}
Asking doubts is an integral part in learning process of any student. Teaching psychology (\& pedagogy) highlights that students should be encouraged to ask doubts and teachers must clarify those doubts in the most comprehensive manner \cite{grade_pred}. It has been an established fact that resolving doubts in an effective manner plays a significant role in improving student's learning outcome. Furthermore, especially in academic settings, asking doubts helps teachers assess student's level of understanding of the concerned topic and promotes active learning. In conventional brick-and-mortar classroom settings, doubt-solving paradigm is dynamic and often asked doubts will be solved instantly. However, beyond the classroom the same teacher is not always available for solving the doubts. To overcome this, students (and parents) opt for a secondary source where the student can learn about the topic and resolve doubts in a timely manner. These secondary sources include physical tuitions or online learning environments ({\it e-}learning platforms). In recent years, wide acceptance of {\it e-}learning platforms, forced by the pandemic (Covid19), has made learning in masses feasible and flexible for addressing the needs of each learner. Especially for doubt-solving at any given time, a few real-time doubt solving online programs were developed. For example Sunali’s Classes etc., is an Indian program which provides 24 hour real-time doubt-solving for school students where the doubts are claimed to be solved in 5 minutes or less \cite{sunali}. Thinkster is a program which helps to visualize the thinking process of a student while solving mathematical problems using AI\cite{thikster}.

\begin{figure}[t] \centering
\includegraphics[height= 3.0in]{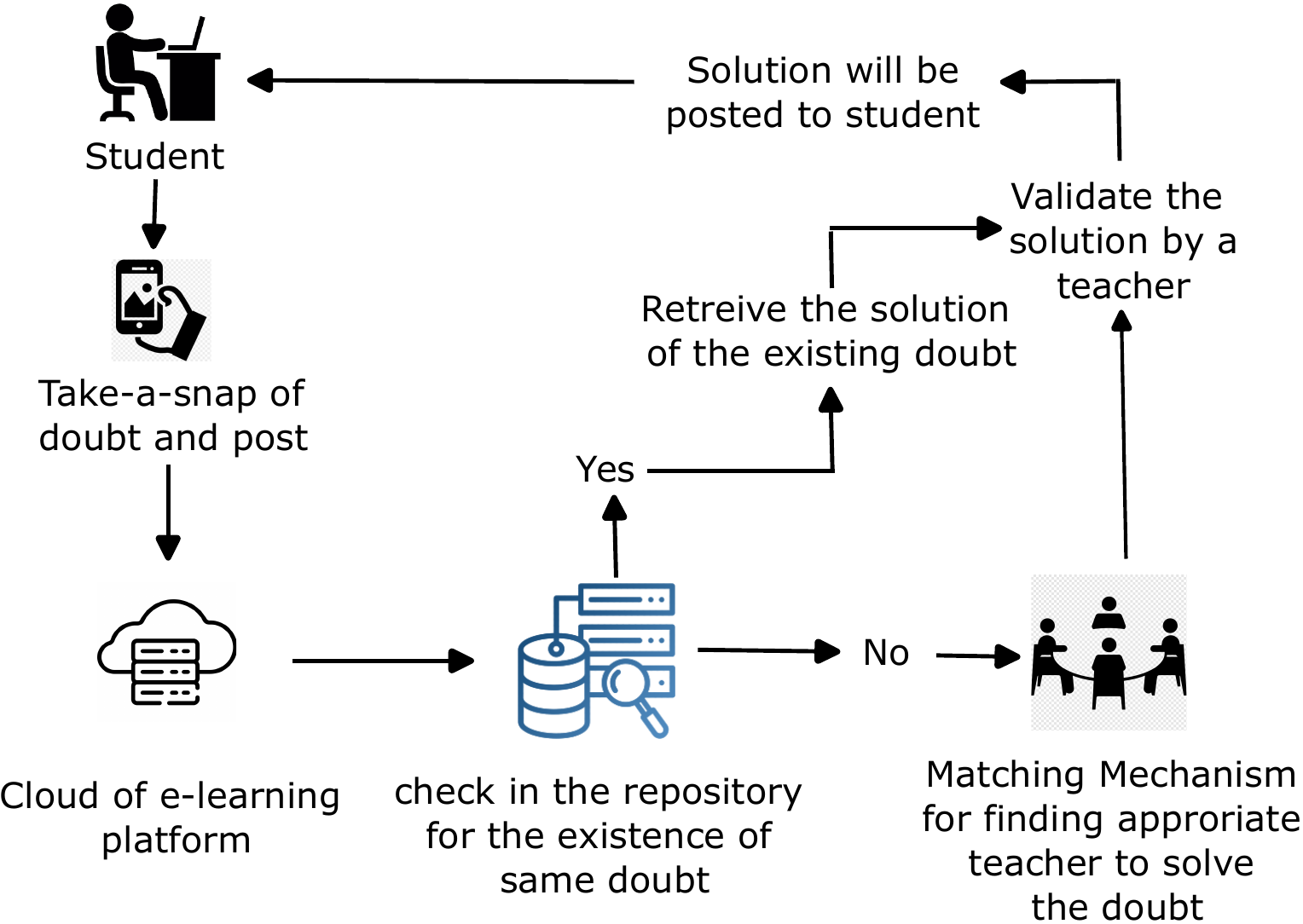}
\caption{General procedure in e-learning platforms.}
\label{Fig:DoubtAskingProcess}
\end{figure}

Most of these e-learning platforms adopted take-a-snap-and-post procedure to ask any doubt, as shown in Fig. \ref{Fig:DoubtAskingProcess}. Much akin to the offline counterparts, these received doubts (images) will be solved by the teachers comprehensively and communicate the same with the student. Each e-learning platform have their own in-house algorithm to identify the appropriate teacher in real-time to solve the incoming doubts. This procedure, however, suffers while solving large volumes of doubts due to the requirement of large man-power (instructors). Otherwise, with less man-power, the time taken to solve each doubt increases significantly. Either of these two options are not appropriate because the resolution time of any doubt defines the quality of interactions between the student \& teacher and hence the engagement with platform. As depicted in Fig. \ref{Fig:DoubtAskingProcess}, to improve on the doubt-resolution time, for frequently asked doubts and/or already solved prior ones, instead of asking the instructor to solve again and post it, the following procedure is considered: 
\begin{itemize}
    \item Check whether or not there exists a match to current doubt in our repository of doubts. 
    \item If a match is found, push the current doubt, matched doubts with the solution, and the confidence of the prior instructor on the solution to the new instructor for validation.
    \item If the current teacher agrees with the existing solution, it will be shared with the student else the doubt will be solved from scratch.
\end{itemize}

Although the aforementioned procedure decreased the doubt-resolution time significantly, its efficiency relies highly on finding the appropriate match from the repository. Generally, the posted doubt can take one of these two forms a) pure text question; b) question with diagram in it. For the former type of doubts, i.e., if the posted doubt has only text, optical character recognition (OCR) is employed to extract the text and then state-of-the-art natural language processing (NLP) techniques like string and semantics matching are used to find the match. Although, OCR is not 100\% accurate, based on the extracted core information the matching is effective. For the later type of doubts, i.e., doubts with little to no text and the diagram in it, OCR is of no use because matching diagram is the essential. To achieve this state-of-the-art computer vision (CV) techniques like ResNet, MobileNet, etc.,  are deployed to match the diagrams and hence find the appropriate match. However, as the students are posting the images, the images tend to contain several adversarial examples, important ones being blurring in some areas, bright and/or dark spots (because of flash usage), pencil and/or pen markings, arbitrary angles which yields rotation and translation of images, and multiple different diagram questions. As a result, regardless of the CV technique, identification of appropriate match is not effective.

Existing deep learning techniques trained for image/object detection models (re-purposed for doubt-matching) are, generally, optimized to achieve the best unbiased estimators for a corresponding label i.e., the conditional expectation given an image. Since the diagrams in the doubts look similar but associated with different doubt(s) and also due to the above mentioned adversarial examples, the conditional expectations on the input image may not be close to the desired estimates. Consequently, these detection models are prone to erroneous matching. This is undesirable since falsely identify matched doubts increases the burden on the validation teacher to go through the match(es), deny them to solve the doubt from scratch, which in turn, increases the doubt resolution time. Although, confidence-interval prediction for each estimate in\cite{conf_interval}, Bayesian neural network approaches in \cite{grade_pred}, and mixture density networks can learn mixture distribution for the understanding of matching variations associated with each estimation. Despite their ability to estimate uncertainties associated with the estimations, these uncertainties are computed after the relationship between input doubt (diagram in it) and the target label has been established. As such, features identified by these models are agnostic to underlying properties that generate these uncertainties, and hence unable to differentiate doubts that resulted in these uncertainties. Furthermore, to learn these uncertainties, a lot of labeled data is necessary for training, which is scarce and expensive. Thus, a model that does not rely on large volumes of labelled data and agnostic to the nuances induced by the adversarial examples associated with doubts is necessary for effective doubt-resolution in e-learning platforms. 

In this work, we propose to frame the task of diagram matching as a representation learning problem that retains the essential information of each diagram despite having variations. Importantly, our approach is fully data-driven in that it does not require any manually generated labels. More specifically, inspired by the recent success of self-supervised learning (SSL) and contrastive learning, we cast the diagram matching  in representation space: the representation of any variation view of an image should be predictive of the representation of another variation view of the same image. The employed SSL technique i.e., BYOL (Bootstrap your own latent space), which directly bootstrap the representations, matches the diagrams by using two neural networks refereed to as online and target network that interact and learn from each other. In this paper, based on theoretical insights of BYOL, we developed a customized augmentation module that mimics the variations in the doubts database. Hereafter, the proposed method is refereed as custom BYOL and the BYOL developed in \cite{BYOL} as default BYOL. The custom BYOL learns a latent space where it learns to project the doubt by retaining essential features of the image ( diagram ), hence grouping the same doubts with variations into a single cluster. The proposed diagram matching engine, first projects the new diagram in the latent space learnt with custom BYOL and then computes the nearest neighbours to identify them as top matches. As the learning mechanism does not require any label information, the proposed approach is robust to outliers, imbalanced data sets. Furthermore, the proposed model is capable of representing a group of never seen samples (out-of-distribution) to a cluster and identify the matches from that newly formulated cluster too, without any manual intervention. Performance of the proposed doubt matching algorithm is evaluated on Vedantu Innovations pvt. ltd. doubt repository. Results demonstrated that the proposed doubt matching engine based doubt matching outperforms its supervised learning counterparts.

This paper is organized as follows: we discuss the related work in Section 2, specify the task definition and describe our self-supervised learning based models in Section 3. In Section 4, we explain the experimental setup, in Section 5 we provided the implementation details. Lastly, we present the evaluation results as well as a detailed analysis.

\section{Related Works}
\subsection{Class Imbalance} The biggest challenge in diagram based search is that, at the time of training we do not have all categories of diagrams available to us and for each category there are hugely varying number of samples. Furthermore, the nature of diagrams would keep on changing as the academic content keeps on evolving year after year. In image classification literature such kind of class imbalance is usually handled by re-balancing the prior distribution through oversampling from the minority class and under-sampling from the majority class. In \cite{SMOTE}, SMOTE is proposed and it applied both these approaches to achieve a superior classification performance. The pitfalls of this approach is that oversampling can lead to overfitting of the model \& downsampling leads to forgoing of useful information. In \cite{Class_balance}, a weighting scheme is introduced for addressing the class imbalance via heavy penalty for the model when it mis-classifies instances from the minority class. 

\subsection{Image retrieval} In \cite{Image_Retrieval}, a large scale image retrieval system called deep local feature or DELF is developed to perform local feature extraction from an image followed by an attention mechanism to focus only on a relevant subset of features for search.   The model is trained with weak supervision and image level labels to extract useful information from query images. All of these approaches are a quick fix towards improving the training of classifiers but it still does not address the problem of dynamic behaviour of number of classes which would require us to retrain the model each time a new class is introduced in our fixed class classification pipeline.

\subsection{Supervised way of learning} 

All \cite{FaceVerification, OneShot, FaceNet}, of these works depend on siamese network architecture that addresses the problem of less training data per class and also handles the problem of varying number of target classes in classification setup. The main objective of this model is to enable us to quantify the similarity between two input data points (diagrams). 

In \cite{FaceVerification}, the authors explain the problem of similarity learning from data in terms of the task of face verification.  They try to learn from a family of functions $G_{W}(X)$ parameterized by W, that captures input patterns in the latent space such that the L1 norm in this compressed space approximates the semantic similarity in the input space. All this is done via a supervised discriminative loss function. They also make use of energy based models that assign low values of energy ($E(X_{1},X_{2})$) to semantically similar pairs of input and higher energy values to semantically dissimilar ones.

In \cite{OneShot}, authors try to perform character recognition by learning a verification model that is able to assign appropriate probabilities to samples from same or different classes. New data points were compared with one sample per novel class and were awarded the class with which they had the highest score. This one shot recognition was based on the idea that once the verification network has learnt an optimum set of weights then it would be powerful enough to extract discriminative features that will allow this framework to generalize to new data as well as new classes from  unknown distributions.

FaceNet in \cite{FaceNet} introduce the concept of hard negative mining in triplet losses for the task of face verification. The network tries to learn from triplets where a negative sample is closer to a given anchor compared to the anchor's positive counterpart. The authors introduce both online (in batch) \& offline (dataset subset) triplet generation methods for faster convergence of the model.

\subsection{Self-supervised works (SSL)}
The Siamese architecture is ideal for our doubt matching engine but curating a large scale labelled dataset to generate appropriate positive pairs is a huge challenge. Therefore we explore SSL mechanisms to avoid this costly labelling step in our work. 

In SSL literature, pretext tasks were the first mechanisms used to train the base encoder models. In this objective a transformation is applied to the input image and the model is forced to predict some aspect of the transformation. The idea behind this approach was that in order to perform well on the given objective, the base model needs to learn about the underlying structure of the entities in an image. Each pretext task aims at modeling some property of the input data distribution. 

Some of the pretext tasks are a) image colorization \cite{Image_Colorization} which aims at understanding different low level features based on the relation of the color between the pixels; b) image rotation prediction \cite{Image_rotation} where the task is to predict the angle by which we rotate the input in order to understand the general ordering of objects in an image; and c) context prediction \cite{Context_Prediction} where a pair of patches is randomly sampled from an image and the goal is to predict the relative position of one with respect to the other.

In PIRL \cite{PIRL}, the problem with these aforementioned works is highlighted and stated that the representations learnt by these methods is covariant with the applied transformation. In reality, applications of various transformations do not modify the semantic properties of an image therefore its representations should be invariant in nature. The authors introduced a mechanism in which they use the pretext task of image jigsaw puzzle solving along with contrastive learning and memory banks for negative samples generation. The final pre-trained model was able to  beat the supervised baseline when it was compared on the downstream task of object detection.

 In \cite{SimCLR}, self supervised based pre-training in linear evaluation protocol with the introduction of a new framework called SimCLR was developed. This framework based on instance discrimination used random cropping \& color jittering to create a positive pair of data points and these points were contrasted with all the other elements in a batch. A modified version of the InfoNCE loss \cite{InfoNCE} that included a temperature scaling parameter, gave the loss function a hardness aware property which forced the model to learn from informative pairs in a batch. Furthermore, a new projection head on top of the base encoder during pre-training phase to improve the quality of information packed in the representations generated by the encoder network is introduced. In the updated version \cite{SimCLRv2} bigger self-supervised models are trained and their label efficiency was demonstrated in the supervised fine tuning steps. 

MoCo \cite{MocoV1}  is defined as an efficient way to perform contrastive based self-supervised learning, where a dynamic queue of representations generated in previous epochs and a momentum encoder are maintained. This method of storing representations allowed to maintain large and consistent dictionaries for performing efficient contrastive learning. In version 2 \cite{MocoV2} the projection head idea of SimCLR along with stronger augmentation functions to achieve improved baseline scores was implemented. 

In \cite{BYOL}, Bootstrap your own latent (BYOL) that depends only on positive pairs for performing contrastive based self-supervised learning is developed. This dependence on positive pairs was crucial because the training dataset curated in this work contains a lot of repetitive samples and the instance discrimination objective could lead to contrasting with data points of similar nature which will eventually lead to learning of incorrect set of features. An asymmetrical architecture that is able to prevent representational collapse and learn features that give state-of-the-art results in ImageNet based on linear evaluation protocol is proposed. The removal of negative sample dependency shows that the model is able to learn features by using small batches and less aggressive augmentations without much loss in performance. Thus, in this work, we chose BYOL to learn the representations of doubts.

\section{The Proposed Doubt Matching Algorithm}
We denote the doubt repository by $ \mathbf{\chi} = \{{\bf x}_1, {\bf x}_2, \cdots, {\bf x}_N\} $, where ${\bf x}_i$ represents $i-$th doubt. Given a doubt ${\bf x}_i \in \mathbf{R}^{m \times m}$, our goal is to find the top-$k$ nearest matches $\hat{{\bf x}}_{i,k} \in \mathbf{R}^{m \times m}$, $k=1, 3, 5$ from the repository $\mathbf{\chi}$. The proposed doubt matching algorithm has two modules: a) Latent space representation and b) matching module, as shown in Fig. \ref{Architecure}. The input doubt provided to the proposed doubt matching engine is first cropped into diagram only by using the YOLO \cite{yolo} architecture. The cropped diagram is then sent to the latent space representation module. In this module, a latent space $f(\cdot)$ where the semantic information between the image and its variants (such as skewed, jittered, croped versions etc.,) are represented as one cluster is learnt, can be given as  
\begin{eqnarray}
    g = \argmin_g \mathbf{E}[||g({\bf z}_i)-{\bf z}`_i||^2_2]
\end{eqnarray}

where ${\bf z}_i = f({\bf x}_i)$ is an arbitrary representation of ${\bf x}_i$, ${\bf z}`_i = f`({\bf x}`_i)$ is an arbitrary representation of ${\bf x}`_i$, and ${\bf x}`_i$ is a variation of ${\bf x}_i$. 

The representation obtained for the given input doubt is then provided to the matching module, as shown in Fig. \ref{Architecure}.  This module identifies the nearest neighbours to learnt representation in the latent space $f$ by

\begin{eqnarray}
    \hat{{\bf x}}_k = \argmin_{{\bf x} \in \chi} \mathbf{E}[||f({\bf x}_i)-f({\bf x})||^2_2].
\end{eqnarray}

\subsection{Latent Space Representation Module}
\begin{figure}[h]\centering 
    \includegraphics[width=110mm]{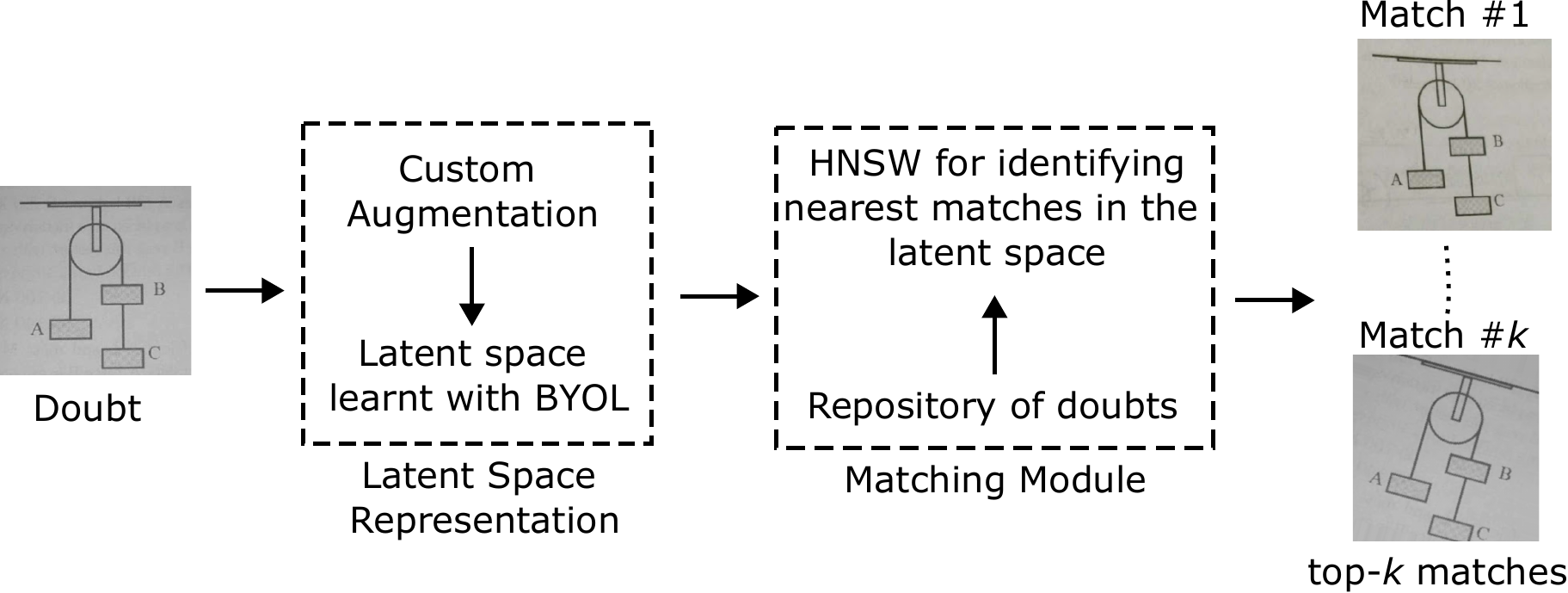}
    \caption{Proposed doubt matching engine.}
    \label{Architecure}
\end{figure}

This module comprises of two main components:  a) custom augmentation and b) latent space representation, as shown in Fig. \ref{Architecure}.

\subsubsection{Custom Augmentation} Augmentation functions are an integral part of SSL paradigm. Existing frameworks include SimCLR \cite{SimCLR}, BYOL \cite{BYOL} employed random cropping followed by color jittering as the augmentation schemes. It has been proved that these augmentations are effective in learning generalized features. However, in \cite{dssl}, it was shown that those augmentation schemes works for ImageNet mainly due to the nature of the data-set i.e., single, object-centric. 

\begin{figure*}[t]\centering 
    \includegraphics[width=\textwidth]{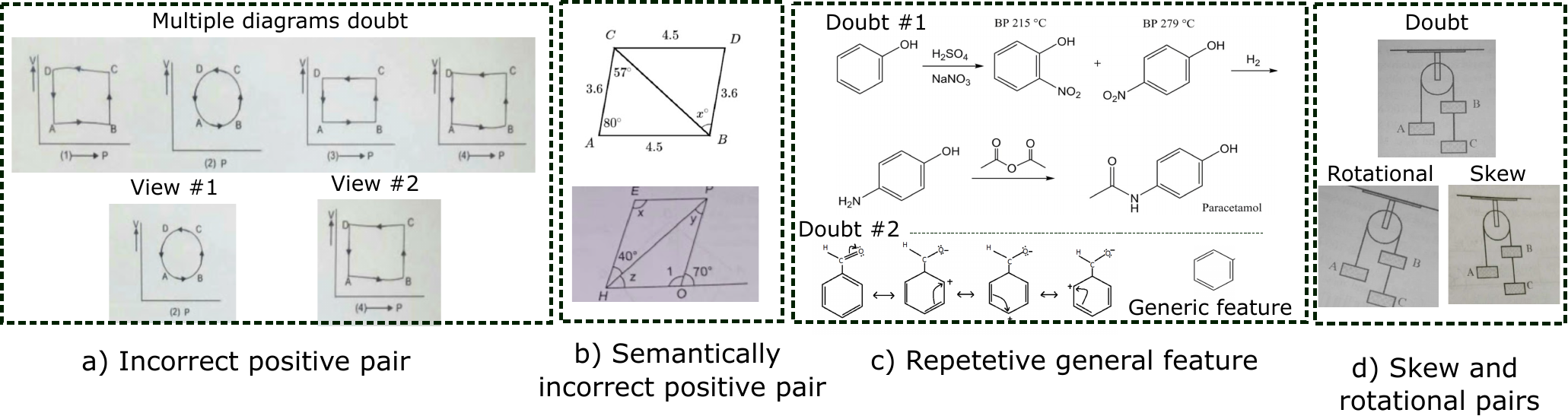}
    \caption{Effect of augmentations in creating positive pairs for training. }
    \label{Augmentations}
\end{figure*}

The images available in the doubts does not possess the characteristics of ImageNet. Furthermore, there are certain features that are quite relevant to the doubt matching only. Thus, when the existing augmentation schemes are applied on the doubt database (detailed in Section 4.1), results showed that random cropping and color jittering are ineffective in formulating a proper positive pair for learning the latent space with BYOL. What follows is a brief discussion of the scenarios that are prevalent in doubt matching task: 
\begin{itemize}
    \item  In case of doubts with multiple diagrams, as shown in Fig. \ref{Augmentations}(a), cropping and matching smaller, sub-diagrams as similar figures as shown in Fig.\ref{Augmentations}(a) View \#1 and View \#2, leads to the formulation of incorrect positive pairs. Furthermore, as shown in Fig. \ref{Augmentations}(b), in some instances, semantically incorrect information due to random flipping (augmentation used in default BYOL) is considered as positive pairs to train with. 
    \item Due to the random cropping, more often general features are highlighted and other useful features are dropped, as shown in Fig. \ref{Augmentations}(c). For both the doubts Doubt $\#1$ and Doubt \#2 depicted in Fig. \ref{Augmentations}(c), the dominating feature is Benzene ring (shown as general feature in Fig. \ref{Augmentations}(c)). As a result of random cropping augmentation, both the doubts are considered as same and thus creates the positive pairs. But, in reality it is a negative pair. 
    \item In some instances, there exists random skew or rotations, as depicted in Fig. \ref{Augmentations}(d). Despite being a positive pair, the noise (like rotations, glittering, pen markings etc), will not be considered with the current augmentation. As a result, despite having no information to gain, the model considers these instances with high information gain and clusters them separately.  
\end{itemize}

Since the semantic relationship of all these pairs is incorrect, the model generalization capabilities are not effective. The importance of modifying the augmentation functions according to the downstream task and its affect on improving the quality of features learnt by SSL model are detailed in \cite{wssl}. Accordingly, in this work, we chose mutual information gain between two views of an augmentation as a metric for selecting that augmentation to construct the positive pairs. When the mutual information between the two views is minimal and the mutual information between each view and the label (as defined by the downstream task) is maximized, then the constructed positive pairs have all the relevant information to learn. Given two views $V_1, \& V_2$ from an augmentation $A$ of an input image ${\bf x}$, then the criteria for selection is 
\begin{eqnarray}
    (V_1*,V_2*) & = & \argmin_{V_1,V_2} \mathrm{I}(V_1,V_2) \\
    \mathrm{I}(V_1; y) & = & \mathrm{I}(V2; y) = \mathrm{I}({\bf x}; y)
    \label{Informationgain}
\end{eqnarray}
where $V_1*, V_2*$ represents final views chosen to create a positive pair, $\mathrm{I}(\cdot, \cdot) $ represents the mutual information gain between the provided views, $y$ represents the label of the image.

As the Augmentation functions in SSL framework determines the set of features to which the model is invariant to and yield similar representation vectors for positive pairs with minimum information gain, as shown in Equ. \ref{Informationgain}, we chose the less aggressive standard random cropping, color jittering, gray scaling, random rotate, channel shuffling, and overlay color mask as the augmentation functions. All the included augmentations are chosen to simulate the potential noise that could be present when a student is posting an image on the search engine. The intensity of random crop and random rotate was limited till it leads to dropping of key features that lead to sub-diagram (false positive) matches. All other augmentation values were handcrafted to resemble the potential noise that could be generated by the users. Analysis presented in automatic evaluation section highlighted the necessity of the chosen augmentation schemes. 

\subsubsection{Latent Space Representation}
\begin{figure}[h]
    \centering
    \includegraphics[width=120mm]{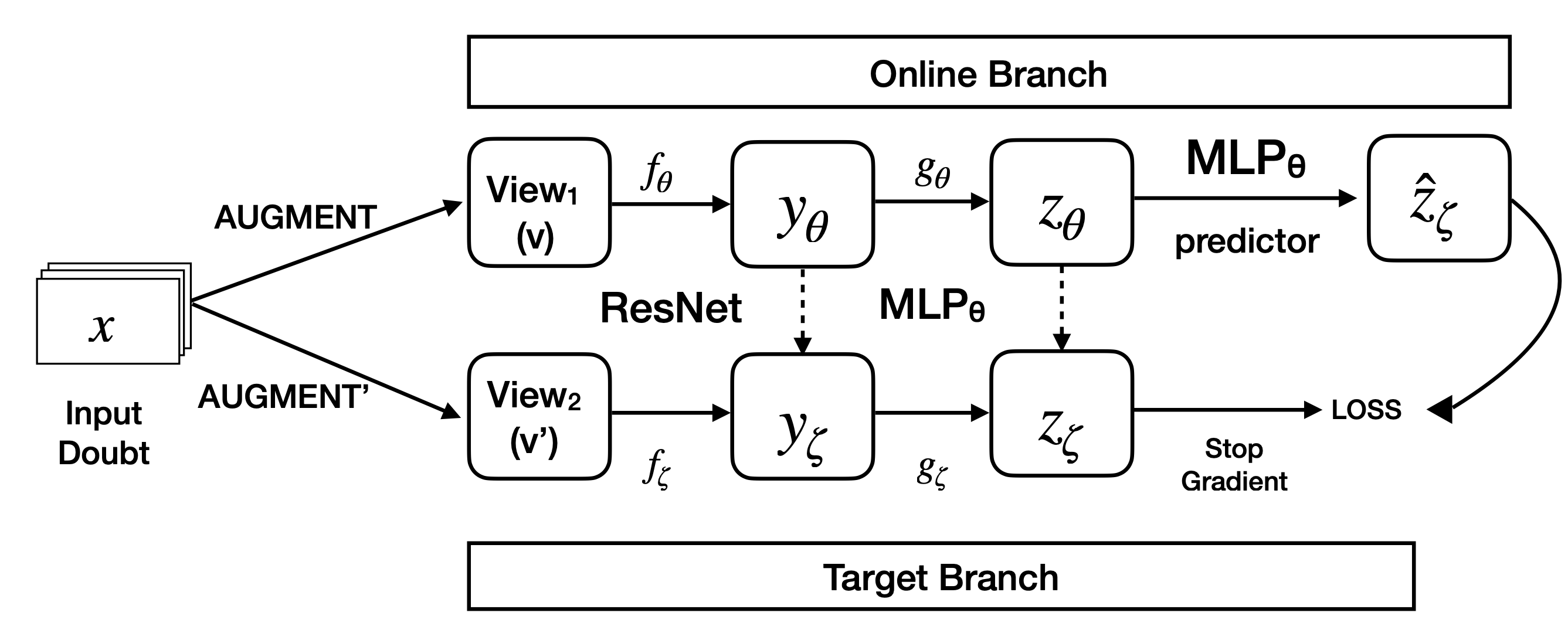}
    \caption{BYOL Architecture \cite{BYOL}} \label{byol}
\end{figure}

The goal of BYOL (Bootstrap Your Own Latent space) is to learn a representation $y_\theta$ which can be use for representing similar doubts as a cluster in an arbitrary latent space. The architecture of BYOL is depicted in Fig. \ref{byol}, and it comprises of two networks i.e., online and target networks. The online network has three stages: an encoder, a projector, and a predictor. The target network has same architecture as online network but uses different set of weights. 

Given a set of doubts $\chi$, a doubt ${\bf x} \in \chi$ and two distributions of image augmentations, two augmented views $v, v'$ of the chosen image will be computed for both online and target networks, respectively, given as:
\begin{eqnarray}
    v = AUGMENT({\bf x})~\&~v' = AUGMENT'({\bf x})
\end{eqnarray}
where $AUGMENT$ and $AUGMENT'$ represents two chosen image augmentations. 

In the online network, the augmented view $v$ is given a base encoder (in this work it is ResNetV2) to provide a representation, can be given as: 
\begin{eqnarray}
    y_\theta = RESNET(v). 
\end{eqnarray}
This representation is projected in to an arbitary space to yield the representation $z_\theta$ with a MLP, can be given as 
\begin{eqnarray}
    z_\theta = MLP(y_\theta). 
\end{eqnarray}
Similar procedure is employed in the target network on the second augmentation view $v'$ to obtain the representation as $y_\zeta$ and projected representation as $z_\zeta$.  

A prediction of $z_\theta$ is computed with a MLP of same dimensions as the projector from the online network named as predictor in Fig.\ref{byol}, can be given as
\begin{eqnarray}
    \hat{z}_\zeta = MLP(z_\theta). 
\end{eqnarray}

Finally, the following mean squared error is computed between the normalized predictions $\hat{z}_\zeta$ and target projections $z_\zeta$, can be given as: 
\begin{eqnarray}
    \mathcal{L}_{\theta, \zeta} = ||\hat{z}_\zeta - z_\zeta||^2_2. 
\end{eqnarray}

The online network is directly updated by the gradient signal generated from the MSE loss function but the target network is updated in a moving average fashion from the parameters of the online network. This ensures that the vectors generated by the target network don't change rapidly \& also the loss value is not directly optimized wrt the parameters of both branches simultaneously, in order to avoid the lowest value of the loss function i.e. constant representations, 
\begin{eqnarray}
    MLP^{*} <- \argmin_{MLP} \mathbb{E} [\parallel MLP(z_{\theta}) - z_\zeta  \parallel_{2}^{2}]; 
\end{eqnarray}
\begin{eqnarray}
    MLP^{*}(z_{\theta}) = \mathbb[z_{\zeta} | z_{\theta}],
\end{eqnarray}
where MLP* is the optimal predictor and the goal is to learn a set of optimal parameters that a generate vector which is close to the vector generated by the target network. This is an iterative process where the online branch captures additional sources of variability about a diagram and that information is passed to the target branch in a moving average style. This passing of information leads to generation of higher quality of representations by the target network, thus forcing the online branch to find more features in the given diagram view to improve further.

\subsection{Matching Module}
Hierarchical navigable small world (HNSW) \cite{HNSW} is employed to identify the nearest matches from the learnt representations. HNSW builds a multi-layered graph structure where each layer represents a proximity graph which is constructed based on our compressed latent space representations of doubts. Elements are populated in the proximity graph based on their closeness to each other. This closenss is computed as the Euclidean distance between the two representations of a given image for supervised settings and cosine similarity for self-supervised settings, can be given as: 
\begin{equation}
   sim(f_\theta({\bf x_1}), f_\theta({\bf x_2}))
\end{equation}
where $sim(.,.)$ represents Euclidean distance or cosine similarity between two representations, ${\bf x_1}$ and ${\bf x_2}$ represents two doubts. 

The layered structure contains nodes with longer edges on the top for fast search \& as we go down, we start to encounter nodes with smaller edges which lead to increase in search accuracy. Search is performed in a greedy manner to find the nearest local minimum vertex to a given query at each layer. This repetition of search at every layer is performed until local minimum is not reached in the lowest layer. The algorithm is able to index millions of vectors in milliseconds without returning too many false positives. For our search engine this algorithm works because both in supervised and self supervised setting, we aim towards arranging similar data points close to each other in the search space based on a distance metric in the pretraining phase.

\section{Experimental Setup}
\subsection{Dataset} As there are no publicly available datasets for developing a doubt matching engine (diagram based search engine), we build a new data set with the doubts repository available at Vedantu Innovations Pvt. Ltd. All the doubts in the repository are manually clustered based on visual similarity. On a whole, 41,371 doubts with images are clustered into 6248 labels. A minimum of three images to maximum of eight images are associated with each cluster. To identify and extract diagrams in those doubts, we developed a diagram detection model based on YOLOv4 \cite{yolo}. This diagram detection model is trained on randomly chosen 10,000 images and validated on 1500 images, achieved a precision of 90\%. Results further highlighted that, the YOLO module always returned a single high confidence box that covered almost all the diagrams in a given image at once. The extracted images and its corresponding cluster labels form the dataset  employed in this work.

\subsection{Performance metrics} The search quality is measured in terms of top-1, -3 \& -5 accuracies i.e. whether the matching candidate to a given doubt is found in these top k results or not. For the clustering quality, we measured precision which in turn computes the number of false positives placed in a given cluster. For the supervised settings (Siamese networks), we employed the triplet loss framework with Euclidean distance as the similarity metric.

\subsection{Baselines} 
In order to select which CNN model will serve best for doubt matching, we compared the search quality on the vectors generated by models trained on ImageNet \cite{imagenet}. As ImageNet dataset comprises of a wide range of images and models trained on it will have better feature extraction capabilities when compared to random baselines. Based on the results, In this study MobileNetV1 \cite{mobilenet} \& ResNet50V2 \cite{resnet} are employed as base encoders for BYOL models. Furthermore, these two models represents the extreme ends of spectrum i.e., latency v/s accuracy trade off in search engines. Performance of these two models are compared in both supervised and self-supervised settings. 

\section{Implementation details}
\subsection{Training Data} In this comparison study, the labelled dataset of 41,371 doubts is divided into training, validation, and test sets with 25,808 samples (belongs to 3385 labels), 5,230 samples (belongs to 847 labels), and 15,564 samples (belongs to 2016 labels), respectively. These splits were created by ensuring that all diagrams belonging to a single cluster are only available in one split, this to prevent any form of data leakage. Furthermore, the test set is carefully selected that same images in test sets are not found in both training and validation sets, so that unbiased performance estimate of our diagram matching model can be reported.

\subsection{Architecture} 
For the supervised training we use the triplet loss framework with euclidean distance as the similarity metric for the siamese network. Both the networks in the supervised setup are initialised with their corresponding ImageNet weights before training. Max pooling operation on the output of base encoder showed a significant advantage in terms of Top-k search score over all the other methods of pooling \& this can be attributed to the fact that if certain key indicator signals exist in the image then max pooling intensifies those signals which allow the base model to generate closer vectors for semantically similar concepts. A double dense layer head was experimented on top of the base network only for fine-tuning, in order to see whether the added non-linearity improves the overall score or not. The added head only degraded the representation even after using a small learning rate to protect the ImageNet weights from the large gradients generated by fine tuned head. 

For the self-supervised training we replicate the BYOL pipeline of online \& target branches to train the base encoder network. The projector \& the predictor networks are updated slightly to meet our training resources. Both networks are 2 layer MLPs of 512 dimensions \& make use of ReLU activation function. Just like the original BYOL, we use batch norm only after the first layer in the MLP. Based on the results of supervised training, again we use global max pooling on the output of the base encoder network.

\subsection{Optimization} 
\begin{figure}[h]
    \centering
    \includegraphics[width=3.5 in]{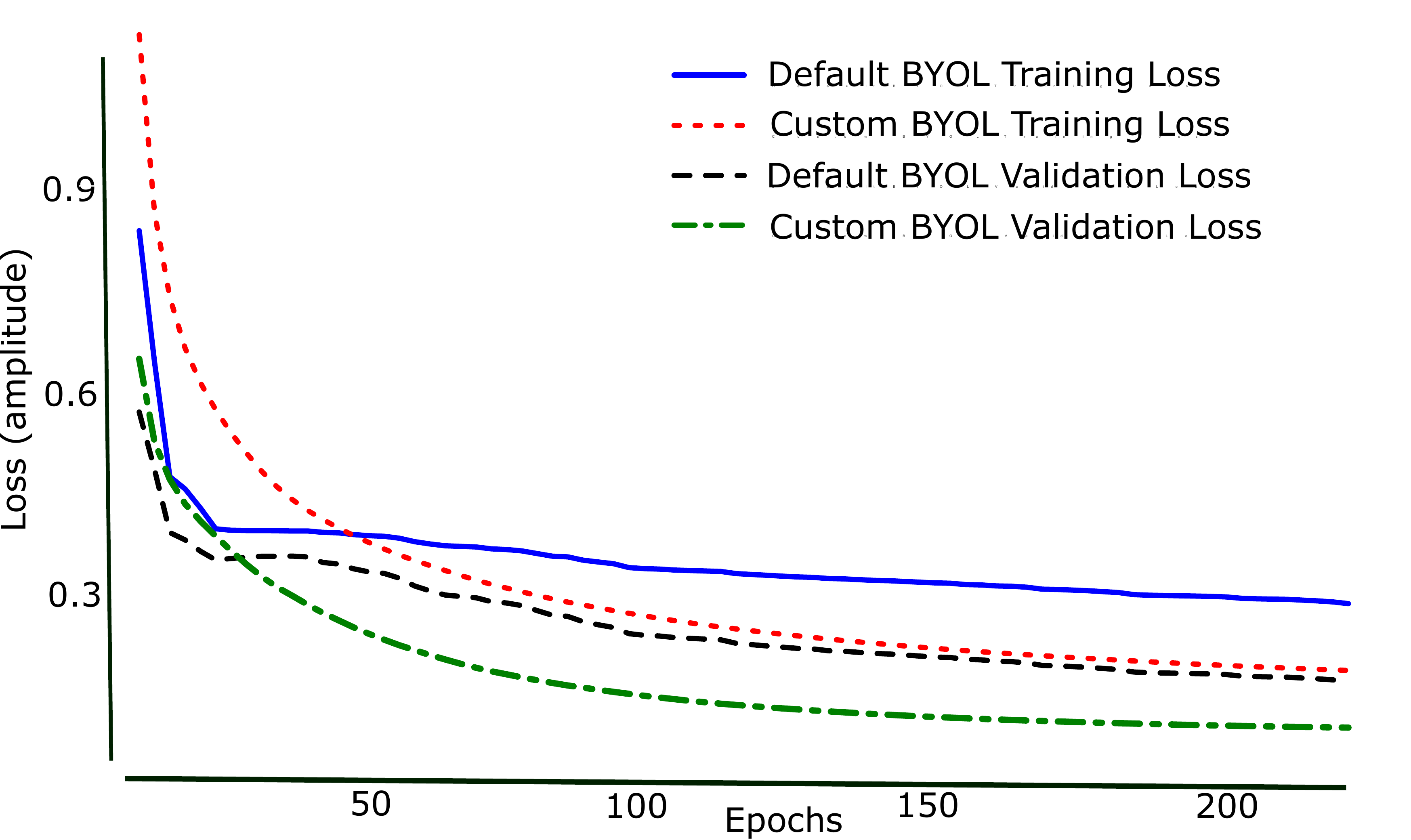}
    \caption{Loss Curves for both custom and defualt BYOL.}
    \label{TrainingandValidationLoss}
\end{figure}

In supervised training regime, both the models i.e. MobileNetV1 \& ResNet50V2 are trained for 200 epochs with a penalty margin of 0.2 in the triplet loss and adam \cite{adam} optimizer with exponential decay of 0.9. For the offline mining mechanism to generate triplets that are used to train the networks, a larger proportion of hard negatives were being generated to improve the overall learning. For MobileNetV1, a batch size of 64 contained 50 hard negatives \& for ResNet50V2 a batch size of 32 contained 20 hard negatives. In the SSL training regime, both the models are trained from scratch for 200 epochs. Adam optimizer and cosine decay with 10 epoch warmup \& a learning rate of 5e-4 is used in the training setup. In SSL a batch size of 128 is supported for MobileNetV1 \& a batch size of 64 is supported for ResNet50V2. All the models are trained on a cloud system with 16gb P100 GPUs. With this setup, training takes approximately 24 hours for a ResNet-50V2 \& 16 hours for a MobileNetV1. In both supervised \& self-supervised settings, the models with best validation loss were saved \& employed to report the performance on test dataset, as shown in Fig. \ref{TrainingandValidationLoss}. As generalization is essential for doubt matching, we evaluated the model performance on validation dataset, which is different from training dataset in terms of the augmentations. In validation dataset, the intensity of each augmentation is reduced. This reduction is to observe whether or not model is learning any useful information for every diagram \& able to match the views accurately. When the validation loss stops improving that indicates that the model is either not learning any useful features or found a trivial way to match training view. Thus, we stop further training and saved the model before saturation of validation loss.

\section{Results and Analysis}
\subsection{Customized Augmentation Layer For Doubt Matching Engine}
Main distinction between defualt BYOL and custom BYOL is the usage of gray scaling, random rotate, channel shuffling, and overlay color mask together with default augmentations random cropping and color jittering. 
Here we ablate the use of augmentations using doubt test set performance on a ResNet50-V2 for 200 epochs. In summary, BYOL achieved a test accuracy of 59.4\%, we gain +6.1\% by adding the custom augmentations on top of the default ones, which yields the final custom BYOL performance of 65.5\%, as tabulated in Table. \ref{tab:lakdc_data}. 

To quantify the effect of each augmentation in performance, we did a comparison study of mutual information gain achieved with the training pairs created with each augmentation. For illustration purpose, mutual information gain (as detailed in Equ. \ref{Informationgain}) , number of positive pairs and negative pairs generated with each augmentation for doubts depicted in Fig. \ref{Augmentations} is tabulated in Table. \ref{mutualinformation}. Results highlight that when compared to cropping (default augmentation), rotation and colour masking (proposed augmentations) are yielding high information gain. This indicates that the proposed augmentations are capable of creating training pairs that are helpful in learning useful representations. Furthermore, because of the random cropping and the existence of rotation/skewness positive pairs are translated to negative pairs and vice versa. In this study, we used a less intense random cropping that retains nearly 70\% of the original image in its smallest crop where as default BYOL's smallest crop retained 8\% of the original image. With the introduction of rotation and less-intense cropping, the positive pairs yielded with chosen augmentations have high mutual information gain and the negative pairs have less. Because of the effective formulation of positive pairs with proposed augmentation, for both the models (i.e., MobileNet and ResNet), custom BYOL yielded higher performance compared to default BYOL. Also the use of aggressive cropping strategies lead to a lot of loss of mutual information gain which is desirable as mentioned in Equ. \ref{Informationgain} but too agressive crops lead to a huge loss in mutual information gain between a given view and its corresponding label ( which we don't know in SSL), thus using less aggressive crops is much more desirable in the given diagram search engine usecase.

\begin{table}[h]
\centering
\caption{Affect of augmentations on mutual information}
\label{mutualinformation}
\begin{tabular}{@{}clcccl@{}}
\toprule
\multicolumn{2}{l}{Augmentation}          & \multicolumn{1}{l}{Cropping} & \multicolumn{1}{l}{Rotation} & \multicolumn{1}{l}{Colour Mask} &  \\ \midrule
\multirow{3}{*}{Original Pair} & Fig .\ref{Augmentations}(a)    & 0.101                        & 0.6104                       & 1.4412                             &  \\ \cmidrule(l){2-6} 
                               & Fig .\ref{Augmentations}(c), Doubt \#1 & 0.0143                       & 0.0294                       & 0.3497                             &  \\ \cmidrule(l){2-6} 
                               & Fig .\ref{Augmentations}(c), Doubt \#2 & 0.0086                       & 0.0576                       & 0.387                              &  \\ \midrule
\multirow{3}{*}{Positive Pair} & Fig .\ref{Augmentations}(a)    & 0.126                        & 0.271                        & 2.1617                             &  \\ \cmidrule(l){2-6} 
                               & Fig .\ref{Augmentations}(c), Doubt \#1 & 0.0188                       & 0.0084                       & 1.4061                             &  \\ \cmidrule(l){2-6} 
                               & Fig .\ref{Augmentations}(c), Doubt \#2 & 0.0301                       & 0.0103                       & 1.4377                             &  \\ \midrule
\multirow{3}{*}{Negative Pair} & Fig .\ref{Augmentations}(a)    & 0.0186                       & 0.01                         & 0.0077                             &  \\ \cmidrule(l){2-6} 
                               & Fig .\ref{Augmentations}(c), Doubt \#1 & 0.0292                       & 0.0067                       & 0.9851                             &  \\ \cmidrule(l){2-6} 
                               & Fig .\ref{Augmentations}(c), Doubt \#2 & 0.0477                       & 0.0132                       & 0.7913                             &  \\ \bottomrule
\end{tabular}
\end{table}

\begin{figure}[h] 
\centering
\includegraphics[width= \textwidth]{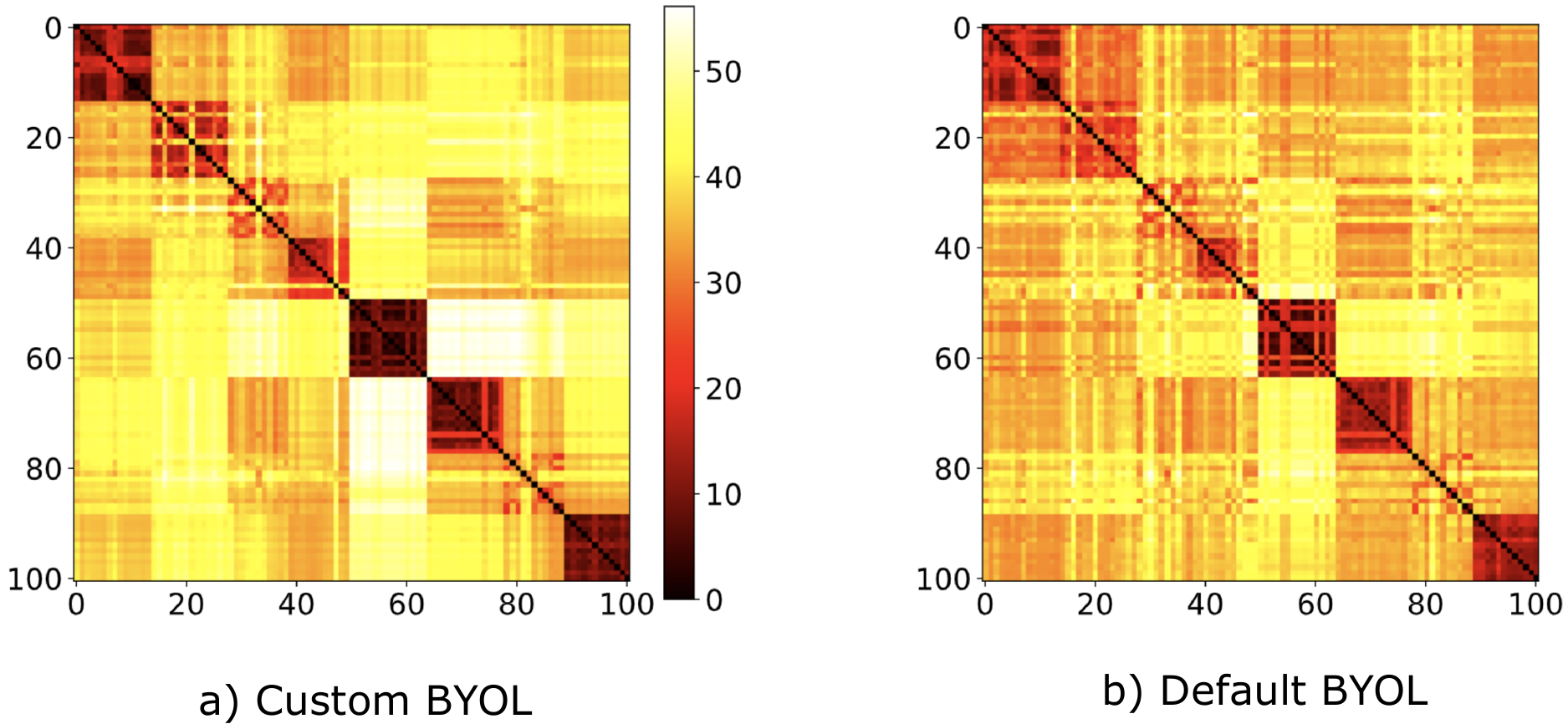}
\caption{Distances between nearest-neighbour representations. Each coloured point in a row represents one of the five nearest neighbours of the representation of that image where the colour indicates the distance between the points.} \label{EucledianPlot}
\end{figure}


\subsection{Effect of Augmentation in the performance of SSL}\label{Augmentationanalaysis}
In order to understand the effect of the custom augmentation on the representations learned by BYOL, we computed the distances between learned representations of closely related classes. The Euclidean distances between nearest-neighbour representations learned by custom  BYOL and default BYOL on doubt dataset, is depicted in Fig. \ref{EucledianPlot}. For this analysis, we randomly picked 8 clusters of doubts that are independent to each other. Each of these 8 clusters have 15 doubts associated with it and ordered contiguously. Each row represents an image and each point in a row represents one of the five nearest neighbours of the representation of that image where color indicates the distance between the image and the nearest neighbour. Representations which align perfectly with the underlying class structure would exhibit a perfect block-diagonal structure; that is their nearest neighbours all belong to the same underlying class. From Fig. \ref{EucledianPlot}(a), it can be seen that with custom BYOL learns representations whose nearest neighbours are closer and exhibit less confusion between labels and super-label than compared to the default BYOL as depicted in Fig. \ref{EucledianPlot}(b).

Comparison analysis conducted on the database to validate the suitability of the proposed augmentation for BYOL compared to its original counterpart is tabulated in Table. \ref{tab:lakdc_data}. Results showed that the proposed augmentation improved BYOL matching accuracy by more than 10\% compared to original augmentation of BYOL, regardless of the base architecture i.e., MobileNetV1 and ResNet50V2. The same trend was observed in top-1, -3, and -5 matches. This further highlights the necessity of customized augmentation based on the nature of the down-stream task.

\begin{table}[h]
\centering
\caption{Performance evaluation for all models in both supervised and self-supervised settings}
\label{tab:lakdc_data}
\centering
\begin{tabular}{ lllll }
\hline
 Model Name &  Top 1 & Top 3 & Top 5\\
 \hline 
 Supervised setting& & & \\ \hline 
 MobileNetV1(ImageNet) & 61.6\% & 69.3\% & 73.4\%\\
 MobileNetV1(Triplet Loss) & 64.1\% & 72.1\% & 75.8\%\\
 MobileNetV1(Default BYOL) & 53.5\% & 60.8\% & 63.4\%\\ 
 MobileNetV1(Custom BYOL) & 64.2\% & 72.5\% & 75.6\%\\
 \hline
 Self-Supervised setting& & & \\ \hline 
 ResNet50V2(ImageNet) & 58.1\% & 65.8\% & 69.2\%\\
 ResNet50V2(Triplet Loss) & 61.3\% & 69.4\% & 72.6\%\\
 ResNet50V2(Default BYOL) & 59.4\% & 66.4\% & 69.3\%\\
 ResNet50V2(Custom BYOL) & 65.5\% & 73.7\% & 77.1\%\\ 
  \hline
\end{tabular}
\end{table}

\subsection{Supervised vs self-supervised} 
Performance of proposed custom BYOL was analyzed in terms of accuracy at top-1, top-3, and top-5 nearest matches to demonstrate the effect of learning appropriate representations on finding the nearest matches in both supervised and self-supervised settings. The accuracy values obtained in both the settings on the testing dataset is listed in Table. \ref{tab:lakdc_data}. Results highlighted that, regardless of the setting, custom augmentation is yielding better performance when compared to the default augmentations.

\begin{figure}[h] 
\centering
\includegraphics[width=100mm]{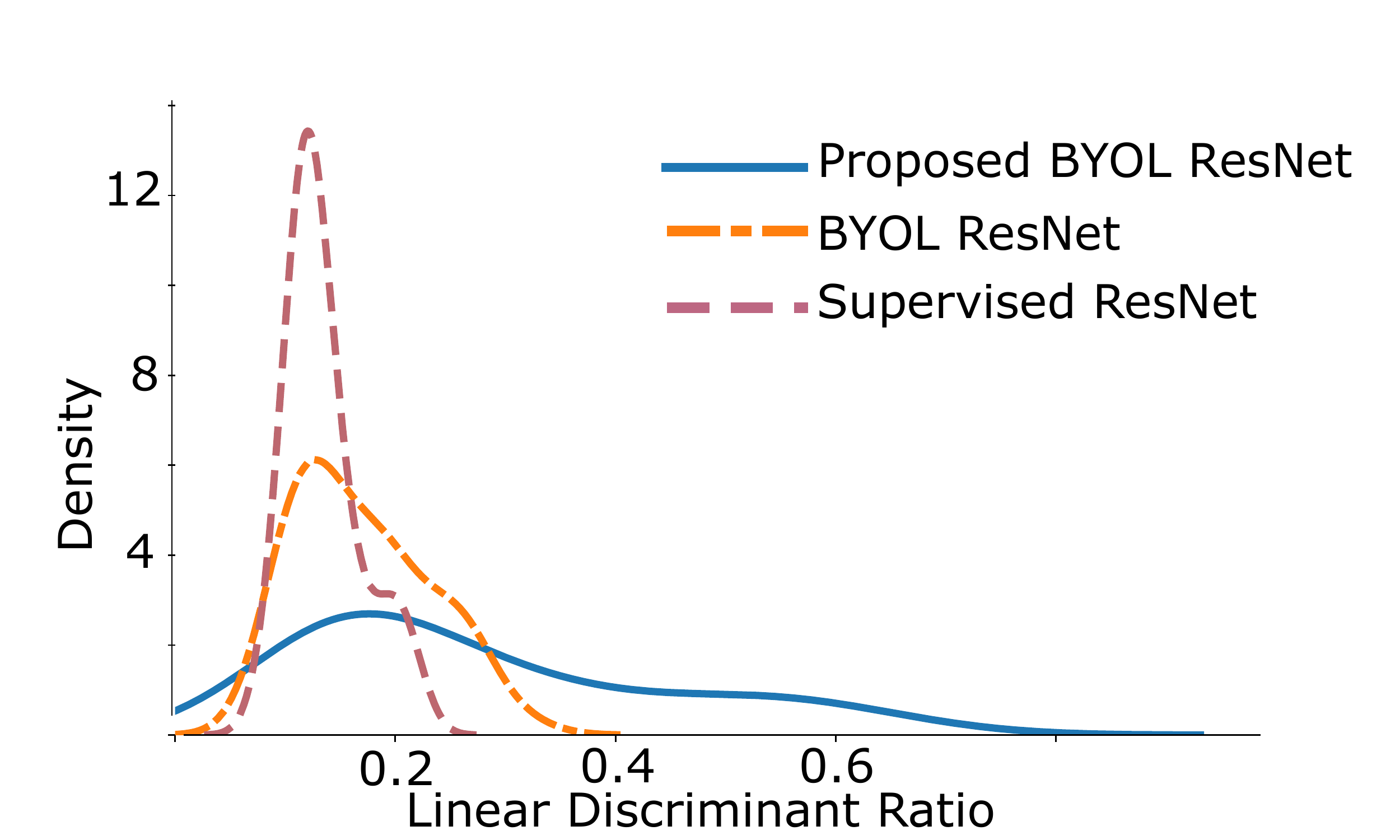}
\caption{Distribution of the linear discriminant ratio: the ratio of between-class distances and within-class distances of embedding computed on the Doubt repository.}
\label{Fig:Schema}
\end{figure}

To quantify the overall structure of the learned latent space, we examine the within- and between-label distances of all labels. The distribution of ratios of between-labels and within-labels  $l_2$ distances of the representation of points in the doubt test dataset learned by BYOL against those learned by its supervised counterparts is depicted in Fig. \ref{Fig:Schema}. A larger ratio implies that the representation is better concentrated within the corresponding labels and better separated between labels and therefore more easily linearly separated (c.f. Fisher’s linear discriminant). From Fig. \ref{Fig:Schema}, it can be seen that for both versions of BYOL's distribution is shifted to the right (i.e., having a higher ratio) compared to the supervised counterpart, suggesting that the representations can be better separated using a linear classifier. When compared between the custom augmented BYOL to the original BYOL, the distribution shifted further right, which indicates the superior representation capabilities with customized augmentations. These empirical results further confirm the theoretical insights of \cite{relicv2} and explain the superior performance of chosen augmentations in Section .\ref{Augmentations}.


\subsection{Human Evaluation} We performed human evaluation studies to measure the quality of matches generated by our system and the humans. Three modalities are considered for evaluation: 1) \# clusters, number of clusters indicates the capability of model in group same images; 2) precision, which measures the effectiveness of model in clustering the same doubts; and 3) noise points, which indicates the samples missed by the model. For evaluation, randomly sampled 10\% of training groups (to use the label information for clustering) is employed to understand how human annotators map different diagrams into one cluster and the model. The clustering is carried out on the vectors in latent space using a combination of UMAP \cite{umap} and HDBSCAN \cite{hdbscan}. Results obtained are tabulated in Table. \ref{tab:humanevaluation}. It can be seen that Custom BYOL and humans have achieved similar performance with 100\% precision. However, BYOL got 90 noise points, which humans grouped with different clusters. This is due to the lack of content or have high discriminating factor. These empirical results highlight that, self-supervised setting is better for doubt matching when compared to both supervised settings and humans.

\begin{table}[h]
\caption{Human Evaluation}
\label{tab:humanevaluation}
\centering
\begin{tabular}{ lllll }
\hline
 Model  &  \# Clusters & Precision & Total & Noise   \\
 \hline
 BYOL & 98 & 100\% & 731 & 90 \\
 \hline
 Manual  & 100 & 100\% & 731 & 0 \\
 \hline
\end{tabular}
\end{table}

\begin{figure}[h] 
\centering
\includegraphics[height= 2.5in]{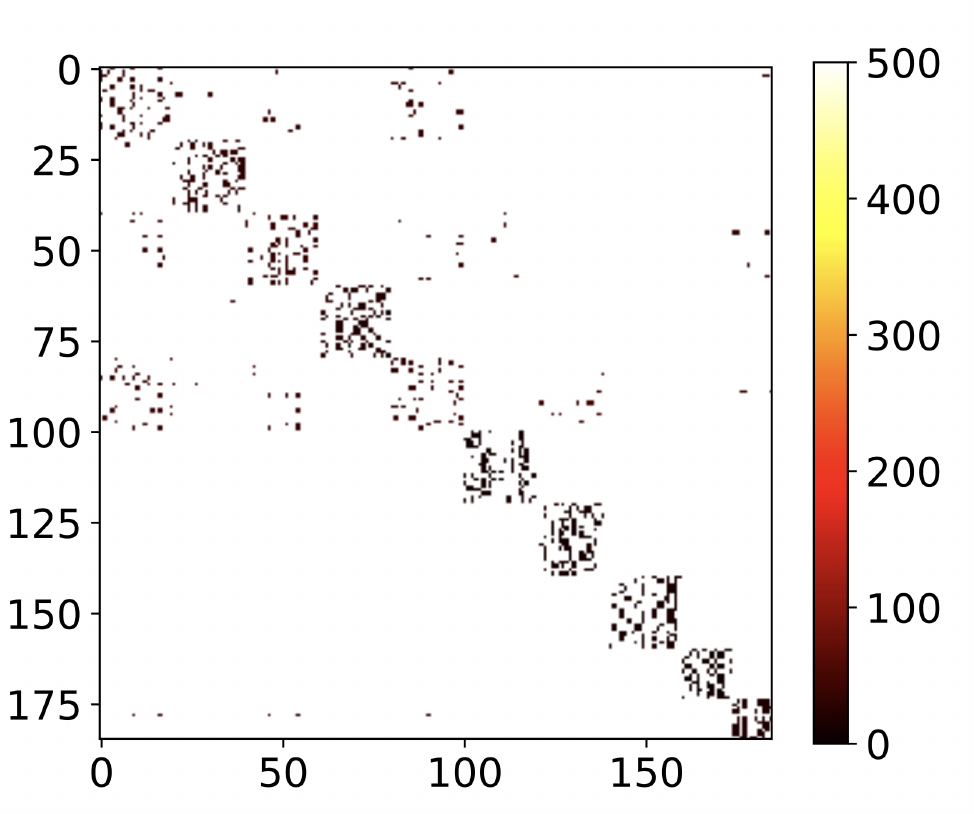}
\caption{Top 5 predictions for out of distribution \& known distribution data points}
\label{Fig:outofdistribution}
\end{figure}

\subsection{Out-of-distribution Samples} 
\begin{figure}[h] 
 \centering
 \includegraphics[height= 1.5in]{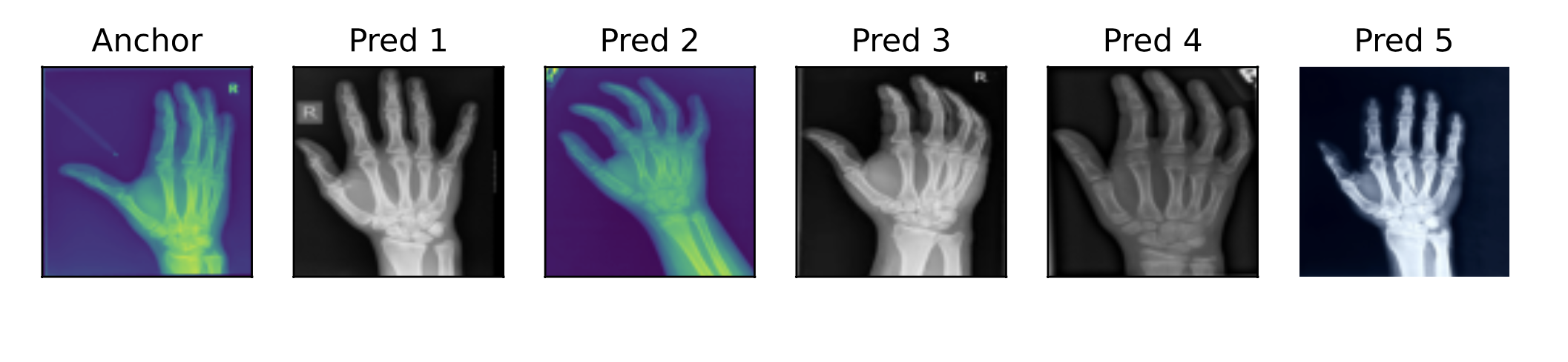}
 \caption{Correct out of distribution prediction}
 \label{corr_oo}
 \end{figure}

\begin{figure}[h] 
 \centering
 \includegraphics[height= 1.5in]{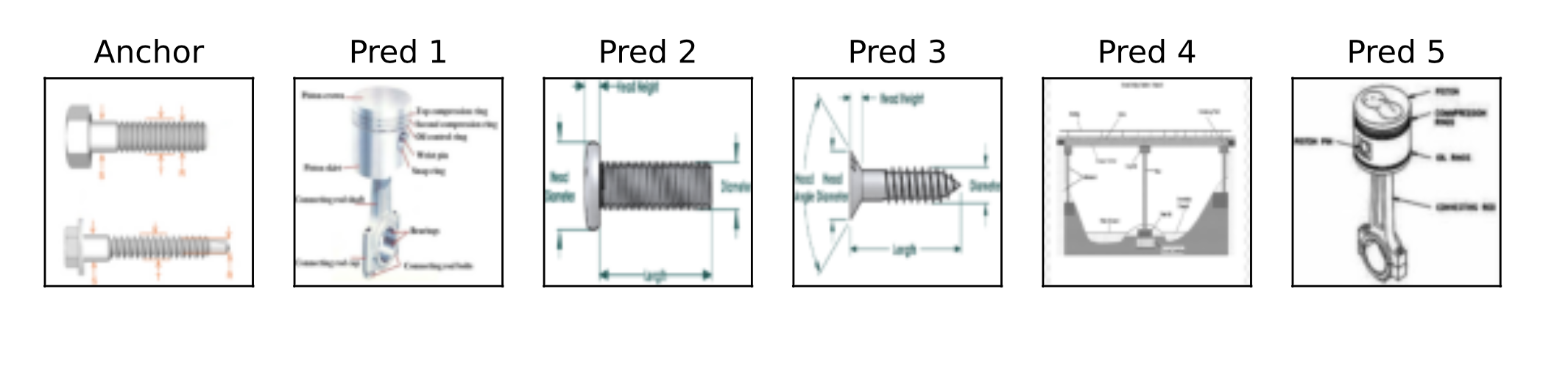}
 \caption{Incorrect out of distribution prediction}
 \label{incor_oo}
 \end{figure}

As custom BYOL performs on par with humans, we evaluated its performance on five never seen clusters acquired from different distribution. Each cluster has five doubts. As detailed in Fig. \ref{EucledianPlot}, the same analysis conducted for the five new clusters together with five randomly chosen clusters from test dataset. From Fig. \ref{Fig:outofdistribution}, it can be seen that despite being a never seen sample, the custom BYOL was able to exhibit less confusion between the labels and super-labels. This implies that, the developed doubt matching engine can organize the new doubts that does not exist in the repository as a new cluster and able to find matches when the relevant doubt is being searched for. Examples of out of distribution queries and predictions can be seen in Fig. \ref{corr_oo} and Fig. \ref{incor_oo}.

\section{Conclusions and Future works}
We have presented self-supervised based approach for automatic doubt matching. We use BYOL technique for the task and investigate the effect of encoding customized augmentations to the default augmentations. Our best model achieves state-of-the-art performance in both automatic evaluations and human evaluations. Here we point out several interesting future research directions. Currently, our  model does not achieve best performance across all variances (noise samples). We would like to explore how to better use the representations learnt to improve the performance of doubt matching of all categories. Besides this, it would also be interesting to consider to incorporate zero-shot learning in our model to further eliminate the human interventions for out-of-distribution samples.  

\bibliographystyle{plain}
\bibliography{references}

\begin{thebibliography}{10}

\bibitem{SMOTE}
N.~V. Chawla, K.~W. Bowyer, L.~O. Hall, and W.~P. Kegelmeyer.
\newblock Smote: Synthetic minority over-sampling technique.
\newblock {\em Journal Of Artificial Intelligence Research, Volume 16, pages
  321-357, 2002}, June 2011.

\bibitem{SimCLR}
Ting Chen, Simon Kornblith, Mohammad Norouzi, and Geoffrey Hinton.
\newblock A simple framework for contrastive learning of visual
  representations.
\newblock February 2020.

\bibitem{SimCLRv2}
Ting Chen, Simon Kornblith, Kevin Swersky, Mohammad Norouzi, and Geoffrey
  Hinton.
\newblock Big self-supervised models are strong semi-supervised learners.
\newblock June 2020.

\bibitem{MocoV2}
Xinlei Chen, Haoqi Fan, Ross Girshick, and Kaiming He.
\newblock Improved baselines with momentum contrastive learning.
\newblock March 2020.

\bibitem{Class_balance}
Yin Cui, Menglin Jia, Tsung-Yi Lin, Yang Song, and Serge Belongie.
\newblock Class-balanced loss based on effective number of samples.
\newblock January 2019.

\bibitem{imagenet}
Jia Deng, Wei Dong, Richard Socher, Li-Jia Li, Kai Li, and Li~Fei-Fei.
\newblock Imagenet: A large-scale hierarchical image database, 2009.

\bibitem{Image_Colorization}
Aditya Deshpande, Jason Rock, and David Forsyth.
\newblock Learning large-scale automatic image colorization.
\newblock pages 567--575, Santiago, Chile, 2015. IEEE.

\bibitem{Context_Prediction}
Carl Doersch, Abhinav Gupta, and Alexei~A. Efros.
\newblock Unsupervised visual representation learning by context prediction.
\newblock May 2015.

\bibitem{Image_rotation}
Spyros Gidaris, Praveer Singh, and Nikos Komodakis.
\newblock Unsupervised representation learning by predicting image rotations.
\newblock March 2018.

\bibitem{BYOL}
Jean-Bastien Grill, Florian Strub, Florent Altché, Corentin Tallec, Pierre~H.
  Richemond, Elena Buchatskaya, Carl Doersch, Bernardo~Avila Pires,
  Zhaohan~Daniel Guo, Mohammad~Gheshlaghi Azar, Bilal Piot, Koray Kavukcuoglu,
  Rémi Munos, and Michal Valko.
\newblock Bootstrap your own latent: A new approach to self-supervised
  learning.
\newblock June 2020.

\bibitem{FaceVerification}
Raia Hadsell, Sumit Chopra, and Yann LeCun.
\newblock Dimensionality reduction by learning an invariant mapping.
\newblock In {\em 2006 {IEEE} Computer Society Conference on Computer Vision
  and Pattern Recognition {(CVPR} 2006), 17-22 June 2006, New York, NY, {USA}},
  pages 1735--1742. {IEEE} Computer Society, 2006.

\bibitem{MocoV1}
Kaiming He, Haoqi Fan, Yuxin Wu, Saining Xie, and Ross Girshick.
\newblock Momentum contrast for unsupervised visual representation learning.
\newblock November 2019.

\bibitem{resnet}
Kaiming He, Xiangyu Zhang, Shaoqing Ren, and Jian Sun.
\newblock Identity mappings in deep residual networks.
\newblock March 2016.

\bibitem{mobilenet}
Andrew~G. Howard, Menglong Zhu, Bo~Chen, Dmitry Kalenichenko, Weijun Wang,
  Tobias Weyand, Marco Andreetto, and Hartwig Adam.
\newblock Mobilenets: Efficient convolutional neural networks for mobile vision
  applications.
\newblock April 2017.

\bibitem{adam}
Diederik~P. Kingma and Jimmy Ba.
\newblock Adam: A method for stochastic optimization.
\newblock December 2014.

\bibitem{OneShot}
Gregory~R. Koch.
\newblock Siamese neural networks for one-shot image recognition.
\newblock 2015.

\bibitem{HNSW}
Yu.~A. Malkov and D.~A. Yashunin.
\newblock Efficient and robust approximate nearest neighbor search using
  hierarchical navigable small world graphs.
\newblock March 2016.

\bibitem{umap}
Leland McInnes and John Healy.
\newblock {UMAP:} uniform manifold approximation and projection for dimension
  reduction.
\newblock {\em CoRR}, abs/1802.03426, 2018.

\bibitem{hdbscan}
Leland McInnes, John Healy, and Steve Astels.
\newblock hdbscan: Hierarchical density based clustering.
\newblock {\em J. Open Source Softw.}, 2(11):205, 2017.

\bibitem{PIRL}
Ishan Misra and Laurens van~der Maaten.
\newblock Self-supervised learning of pretext-invariant representations.
\newblock December 2019.

\bibitem{grade_pred}
Kelvin H.~R. Ng, S.~Tatinati, and Andy W.~H. Khong.
\newblock Grade prediction from multi-valued click-stream traces via
  bayesian-regularized deep neural networks.
\newblock {\em IEEE Transactions on Signal Processing}, 69:1477--1491, 2021.

\bibitem{Image_Retrieval}
Hyeonwoo Noh, Andre Araujo, Jack Sim, Tobias Weyand, and Bohyung Han.
\newblock Large-scale image retrieval with attentive deep local features.
\newblock December 2016.

\bibitem{dssl}
Senthil Purushwalkam and Abhinav Gupta.
\newblock Demystifying contrastive self-supervised learning: Invariances,
  augmentations and dataset biases.
\newblock July 2020.

\bibitem{FaceNet}
Florian Schroff, Dmitry Kalenichenko, and James Philbin.
\newblock Facenet: A unified embedding for face recognition and clustering.
\newblock March 2015.

\bibitem{sunali}
Sunali.
\newblock Sunali's classes.
\newblock {\em https://sunalisclasses.com/}.

\bibitem{conf_interval}
Sivanagaraja Tatinati, Yubo Wang, and Andy W.~H. Khong.
\newblock Hybrid method based on random convolution nodes for short-term wind
  speed forecasting.
\newblock {\em IEEE Transactions on Industrial Informatics}, 18:7019--7029,
  2022.

\bibitem{thikster}
Thinkster.
\newblock Thinkster.
\newblock {\em https://hellothinkster.com/}.

\bibitem{wssl}
Yonglong Tian, Chen Sun, Ben Poole, Dilip Krishnan, Cordelia Schmid, and
  Phillip Isola.
\newblock What makes for good views for contrastive learning?
\newblock May 2020.

\bibitem{relicv2}
Nenad Tomasev, Ioana Bica, Brian McWilliams, Lars Buesing, Razvan Pascanu,
  Charles Blundell, and Jovana Mitrovic.
\newblock Pushing the limits of self-supervised resnets: Can we outperform
  supervised learning without labels on imagenet?
\newblock January 2022.

\bibitem{InfoNCE}
Aaron van~den Oord, Yazhe Li, and Oriol Vinyals.
\newblock Representation learning with contrastive predictive coding.
\newblock July 2018.

\bibitem{yolo}
Chien-Yao Wang, Alexey Bochkovskiy, and Hong-Yuan~Mark Liao.
\newblock Scaled-yolov4: Scaling cross stage partial network.
\newblock November 2020.

\end{thebibliography}
\end{document}